\documentclass[10pt,twocolumn,letterpaper]{article}
\usepackage[pagenumbers]{cvpr}

\usepackage{stfloats}

\usepackage{currfile} 

\usepackage{caption} 

\usepackage{multirow}

\usepackage{xcolor}
\AtBeginDocument{%
    \setlength{\abovedisplayskip}{6pt plus 2pt minus 2pt}%
    \setlength{\belowdisplayskip}{6pt plus 2pt minus 2pt}%
    \setlength{\belowdisplayshortskip}{4pt plus 2pt minus 2pt}%
}

\definecolor{cvprblue}{rgb}{0.21,0.49,0.74}
\definecolor{lightblue}{RGB}{100,100,255}

\usepackage[breaklinks,colorlinks,allcolors=cvprblue]{hyperref}

\title{ReVAgent: Multi-Agent Reasoning-Intensive Video Generation}
\author{%
    Yehui Yang\textsuperscript{1,2}\quad
    Jing Fan\textsuperscript{3}\quad
    Jun Hu\textsuperscript{2}\quad
    Shangyao Chen\textsuperscript{2}\\[3pt]
    Jinhua Bian\textsuperscript{2}\quad
    Song Wang\textsuperscript{2}\quad
    Ruize Han\textsuperscript{2}\thanks{Corresponding author.}\\[6pt]
    {\normalsize\textsuperscript{1}Shenzhen University, Shenzhen, China}\\
    {\normalsize\textsuperscript{2}Shenzhen University of Advanced Technology, Shenzhen, China}\\
    {\normalsize\textsuperscript{3}Yunnan Minzu University, Kunming, China}%
}
\date{}
\hypersetup{%
    pdftitle={ReVAgent: Multi-Agent Reasoning-Intensive Video Generation},
    pdfauthor={Yehui Yang; Jing Fan; Jun Hu; Shangyao Chen; Jinhua Bian; Song Wang; Ruize Han},
    pdfsubject={arXiv preprint},
    pdfkeywords={instructional video generation, multi-agent systems, reasoning, visual explanations}%
}

\newcommand{\arxivwithoutappendix}{}
\begin{document}
\maketitle
\begingroup
\renewcommand{\endabstract}{\par\addvspace{12pt}}
\begin{abstract}
A central challenge in generating instructional videos for reasoning-intensive
topics is to show how and why the state changes at each step. Even when an
animation renders successfully, the depicted object relationships and state
transitions can conflict with the explanation, making the reasoning difficult
to follow. We present ReVAgent, a multi-agent framework that solves
a problem, generates keyframes for its explanation, and synthesizes an
instructional video. The agents share the execution record of a solution
program, reuse intermediate results and constructions in the scenes, and check
rendered keyframes to guide revisions to the solution or the scenes. We also
introduce a multidimensional rubric for assessing whether generated videos
present the problem clearly, organize the steps coherently, and use visuals to
explain the reasoning. We calibrate a multimodal judge using human ratings and
apply it to instructional videos for mathematics and algorithms. In the
evaluated samples, ReVAgent presents multi-step reasoning coherently and
uses diagrams to show how key operations lead to subsequent states.
\end{abstract}
\endgroup

\section{Introduction}
\label{sec:introduction}

\begin{figure*}[!t]
    \centering
    \includegraphics[width=0.9\textwidth]{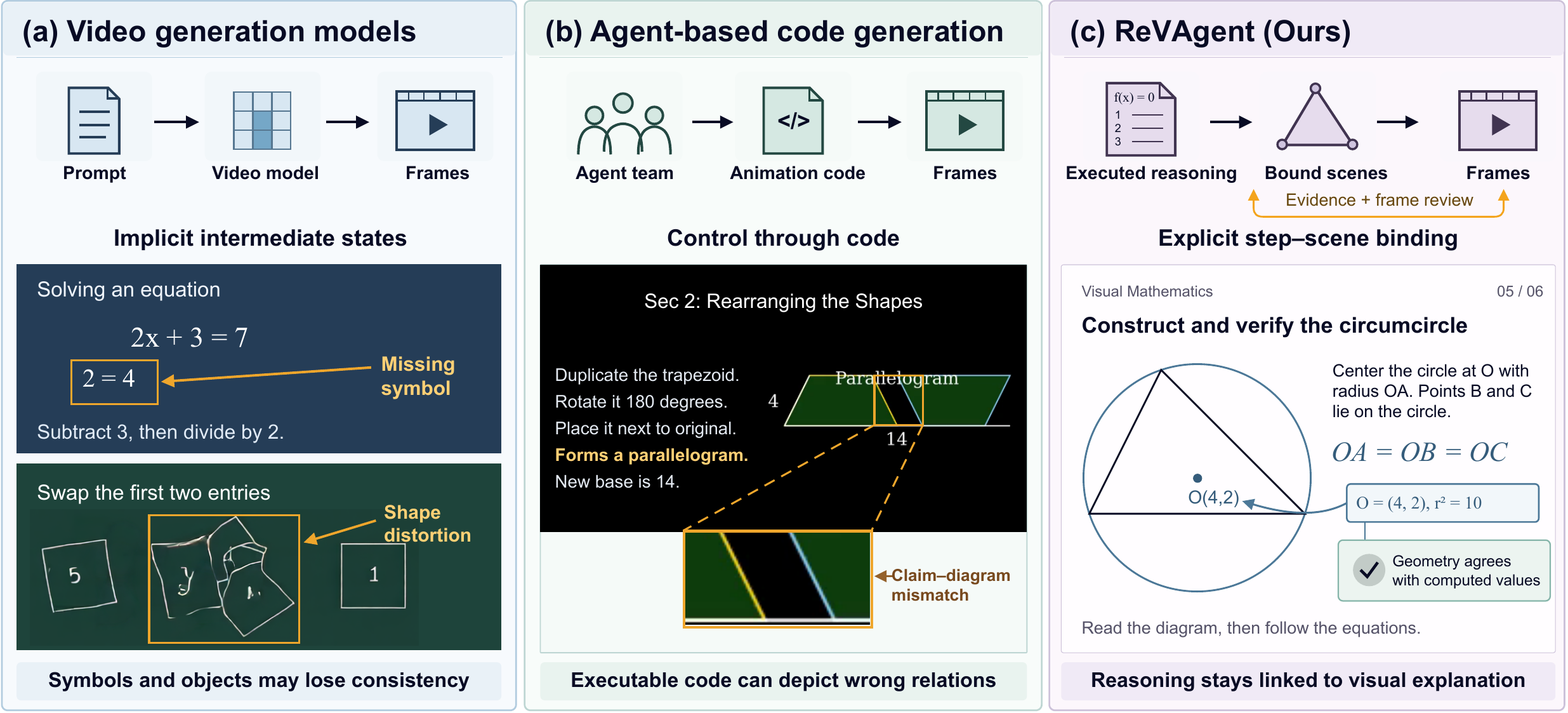} \vspace{-5pt}
    \caption{\textbf{Approaches and challenges in logically coherent educational video generation.}
    (a) Conventional video generation models synthesize frames directly and can
    omit symbols or distort objects in intermediate frames.
    (b) Code-driven agents control visuals through animation scripts, but the
    rendered relations between objects can conflict with the explanation even
    when the scripts execute successfully.
    (c) ReVAgent uses a three-stage pipeline from a textual solution to
    keyframes and then to video. Its agents reuse solution outputs to construct
    scenes, check the resulting keyframes, and provide feedback for revision to
    align formulas, graphics, and their changes with the solution steps.}
    \label{fig:motivation}
    \vspace{-5pt}
\end{figure*}

Video generation has advanced rapidly in recent years, yet generating \emph{reasoning-intensive videos} remains challenging. For example, generating logically coherent educational videos for mathematics and algorithms requires turning solution steps and their rationale into visual explanations that help learners understand how a solution is reached step by step. Let's consider Kruskal's minimum spanning tree algorithm. At each step, the video should preserve the input graph and edge weights, distinguish accepted edges, and show a new edge being accepted only if it joins two different components. Such logical coherence requires each scene's quantities, symbols, and object relations to match the corresponding solution step, and changes between scenes to reflect the computations, constructions, or algorithmic operations being explained.

On one hand, general-purpose video generation models~\cite{gao2025seedance,wan2025}
can synthesize coherent videos from text or reference
images. In videos that teach mathematics or algorithms, however, numbers,
operators, and geometric relations carry information about the solution, so
an error in a single visual element can change the meaning of the explanation.
As Fig.~\ref{fig:motivation}(a) illustrates, even with reference images for the
first and last frames, video generation models~\cite{jiang2025vace} can omit mathematical
symbols or deform objects in intermediate frames. Such errors can cause the visuals to
diverge from the corresponding solution steps, making it difficult for these
models to reliably depict multi-step reasoning in instructional videos.

On the other hand, code-driven approaches offer a promising way to keep symbols and graphics accurate throughout a solution while providing explicit control over what is shown and how it changes. These approaches provide this control through programmatic rendering~\cite{wei2025diagramagent}, using animation tools such as Manim~\cite{coluci2022manim} to draw formulas and graphics from explicit specifications. Advances in large language models (LLMs) enable agents in Code2Video~\cite{chen2026code2video} and TheoremExplainAgent~\cite{ku2025theoremexplainagent} to plan explanations and write executable animation scripts~\cite{liu2025preacher}. Even when scripts run successfully, rendered object relations may conflict with a solution step~\cite{ku2025theoremexplainagent,lopez2026sga}. In Fig.~\ref{fig:motivation}(b), two trapezoids described as forming a parallelogram still leave a gap where they should meet. Even with a correct solution, agents must reimplement its object relations and state changes in animation code. This additional translation makes visual control dependent on code implementation, limiting how fully agents can use their reasoning to organize and adapt visual explanations.

To advance reasoning-intensive video generation with logically coherent visual explanations, we propose ReVAgent, a multi-agent framework for logically coherent educational video generation.
ReVAgent stores the intermediate states and mathematical constructions
produced by the solution program in a shared record and uses them as the basis
for scene construction (Fig.~\ref{fig:motivation}(c)). The visual agent directly
reuses these results to plan the explanation order, organize formulas and
graphics, and choose what to highlight, reducing the need to reimplement
mathematical relations in animation code. A unified renderer converts scene
specifications into video, providing explicit control over symbols, graphics,
and their changes. The verification agent checks the rendered keyframes against
the original problem and solution record to assess their agreement with the
corresponding reasoning steps, and routes detected issues to the responsible
agent for revision. Figure~\ref{fig:output_examples} shows selected frames from
ReVAgent videos on the shoelace formula and Kruskal's algorithm.

\begingroup
\setlength{\parskip}{0pt}
We evaluate ReVAgent on mathematical and algorithmic tasks by comparing
it with state-of-the-art general-purpose and code-driven video generation methods. 
On the reported video subsets, ReVAgent obtains competitive scores on video quality and semantic metrics and
ranks highest among the compared generation methods on all six instructional
dimensions assessed by the human-calibrated judge.
Our main contributions are as follows.

\begin{enumerate}[topsep=0pt,partopsep=0pt]
    \item \textbf{Staged, controllable generation for reasoning-intensive videos.}
    We propose a staged framework from solutions through keyframes to video.
    Structured scene specifications and a unified renderer explicitly control
    formulas, graphics, and their changes at each solution step, reducing symbol
    omissions and object distortions in direct video synthesis.

    \item \textbf{Reasoning-visual coordination through shared solution records.}
    Shared solution records connect solving, scene planning, and frame
    verification. The visual agent directly reuses computed results and geometric
    constructions, reducing the need to reimplement mathematical relations in
    animation code. Frame verification routes detected issues to the responsible
    agents for revision, improving consistency between visuals and solution steps.

    \item \textbf{Multidimensional instructional evaluation.} We define
    criteria including step coherence, reasoning clarity, and visual helpfulness,
    and calibrate a multimodal judge with human ratings.
\end{enumerate}
\endgroup

\begin{figure*}[!t]
    \centering
    \input{figures/motivation/output_examples}
    \caption{\textbf{Selected frames from ReVAgent videos.}
    (a) In the shoelace-formula video, the vertex coordinates are preserved and
    the edge used in the current calculation is highlighted.
    (b) The Kruskal video highlights the edge that merges two components and
    shows the final minimum spanning tree with its total weight.}
    \label{fig:output_examples}
    \vspace{-3pt}
\end{figure*}

\section{Related Work}
\label{sec:related_work}

\paragraph{General-purpose video generation.}
Large-scale video generation models have extended synthesis to longer clips and
multi-shot sequences~\cite{guo2025lct,kara2025shotadapter,meng2026holocine}, with
text or reference images guiding the output~\cite{lin2025stiv,zhang2026stage}.
Sora~\cite{brooks2024sora} demonstrates generation across a range of durations,
resolutions, and aspect ratios, and Seedance 1.0~\cite{gao2025seedance} supports
multi-shot generation from text or images. Open-weight models such as
Wan~\cite{wan2025} and CogVideoX~\cite{yang2025cogvideox} provide publicly
available implementations for comparative study~\cite{yang2026longlive,huang2025selfforcing}. Mathematical and algorithmic
instruction also requires that formulas, relations between depicted objects,
and state changes across scenes agree with the corresponding solution steps.
This requirement motivates code-driven approaches that specify visual elements
and their changes explicitly~\cite{chen2026code2video}.

\vspace{-8pt}
\paragraph{Code-driven educational video generation.}
Animation libraries such as Manim~\cite{coluci2022manim} let authors write
scripts that specify the layout, transformations, and timing of formulas and
graphics. As large language models (LLMs) gained the ability to write these
scripts from instructional content, agent-based systems began to divide the
work into planning, code generation, and revision after rendering.
TheoremExplainAgent~\cite{ku2025theoremexplainagent} plans theorem explanations
and implements them as Manim animations. Code2Video~\cite{chen2026code2video}
turns lecture content into storyboards and uses visual feedback to refine
the rendered layout. VisualEDU~\cite{chen2025visualedu} uses code and visual
feedback to generate step-by-step explanations of mathematical problems.
LASEV~\cite{yan2026lasev} coordinates solution, illustration, and narration
agents, with the illustration agent generating code from a reviewed textual
solution. For algorithm visualization, ALGOGEN~\cite{liaokunpeng2026algogen}
renders execution traces emitted by generated trackers. SGA~\cite{lopez2026sga}
extracts scene graphs from animation code to detect spatial conflicts.
ReVAgent links instructional scenes to executed solution steps, directly
reuses selected geometric constructions produced by the solver, and uses the
same solution evidence to review the rendered keyframes.

\section{Method}
\label{sec:method}

\subsection{Problem Formulation and Overview}
\label{sec:overview}

Given a problem $q$, our goal is to generate an educational video $V$ whose explanations, formulas, and graphics follow the steps of its solution. The input $q$ includes the problem statement, its conditions and numerical data, and any provided diagrams or presentation requirements.

\begin{figure*}[t]
    \centering
    \includegraphics[width=\textwidth]{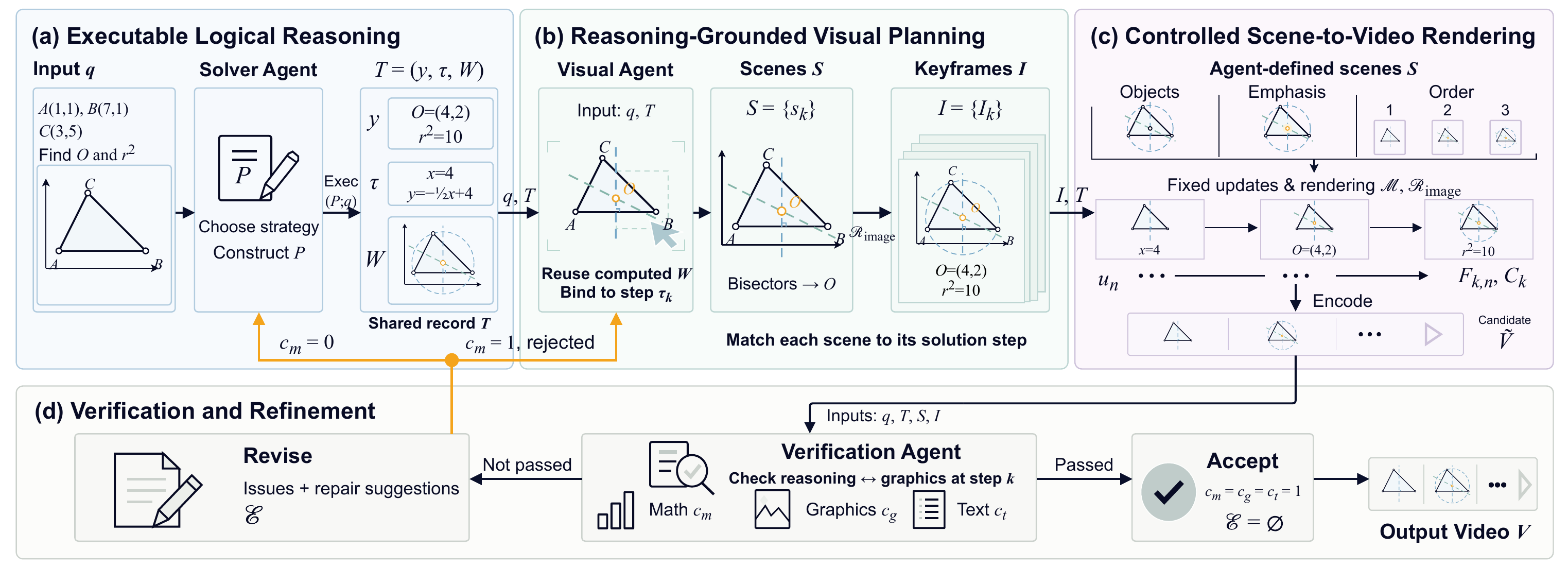}\par
    \vspace{-5pt}
    \caption{\textbf{ReVAgent framework illustrated with a circumcircle problem.}
    (a) Given $A(1,1)$, $B(7,1)$, and $C(3,5)$, the solver computes $x=4$ and $y=-\tfrac12x+4$ as the perpendicular bisectors of $AB$ and $AC$, respectively. They intersect at $O=(4,2)$, with $OA^2=OB^2=OC^2=10$. The execution record $T=(y,\tau,W)$ stores the answer $y=(4,2,10)$ and the constructions.
    (b) The visual agent reuses the stored bisectors and circle to illustrate how the circumcenter is found.
    (c) The renderer composes the scenes into a candidate video.
    (d) The verifier checks the rendered keyframes against the problem, execution record, and scene specifications before acceptance and routes defects for repair.}
    \label{fig:framework}
    \vspace{-9pt}
\end{figure*}

Figure~\ref{fig:framework} summarizes ReVAgent. In \emph{Problem Solving}, a solver agent generates a program and executes it to obtain a solution record $T$. In \emph{Visual Explanation}, a visual agent uses $T$ to plan an ordered scene sequence $S$, from which a fixed renderer produces keyframes $I$. In \emph{Video Synthesis}, the renderer composes the scenes into a candidate video $\widetilde V$.
\begin{equation}
    q \xrightarrow[\operatorname{Exec}]{\mathcal{A}_{\mathrm{solve}}} T
    \xrightarrow[\mathcal{R}_{\mathrm{image}}]{\mathcal{A}_{\mathrm{visual}}} (S,I)
    \xrightarrow[\operatorname{Encode}]{\mathcal{M}} \widetilde V.
    \label{eq:pipeline}
\end{equation}
The visual agent reuses computed graphics and coordinates; the verifier checks frames against the corresponding solution steps, enabling reasoning--visual alignment across agents. In \emph{Verification and Refinement}, the verifier reviews each candidate's keyframes $I$ alongside $q$, $T$, and $S$. A passing candidate becomes $V$. When the verifier identifies defects, it sends them to the agent responsible for revision.

\subsection{Executable Logical Reasoning}
\label{sec:text_stage}

The solver agent chooses a strategy for $q$ and expresses it as a program $P$. The executor runs $P$ on the numerical data supplied in $q$ and collects the returned solution record $T$.
\begin{equation}
    P=\mathcal{A}_{\mathrm{solve}}(q),\qquad
    T=(y,\tau,W)=\operatorname{Exec}(P;q).
    \label{eq:text_execution}
\end{equation}
The answer $y$ contains the requested quantities. The trace $\tau$ records step identifiers, explanations, and intermediate computational states. The evidence $W$ stores computed checks and any geometric constructions returned by the program, including their coordinates and primitive parameters. The visual agent can reuse these constructions directly. The verifier assesses the mathematical validity of the answer, step claims, and constructions against $q$.

\subsection{Reasoning-Grounded Visual Planning}
\label{sec:image_stage}

The visual agent organizes the explanation into an ordered sequence $S=(s_1,\ldots,s_K)$.
\begin{equation}
    s_k=(b_k,e_k,\phi_k,\mathcal{O}_k,\mathcal{H}_k),
    \qquad \mathcal{H}_k\subseteq\mathcal{O}_k,
    \label{eq:scene_representation}
\end{equation}
Here, $b_k$ identifies the corresponding step in $\tau$, $e_k$ and $\phi_k$ specify the instructional text and formula, and $\mathcal{O}_k$ and $\mathcal{H}_k$ denote the visible and highlighted objects, respectively. The step identifier lets the verifier compare the scene with its corresponding computational state.

The visual agent can select constructions from the execution record and add explanatory objects.
\begin{equation}
    \mathcal{O}_k=\mathcal{O}_k^{\mathrm{exec}}\cup\mathcal{O}_k^{\mathrm{gen}},
    \qquad \mathcal{O}_k^{\mathrm{exec}}\subseteq\mathcal{O}(W).
    \label{eq:scene_objects}
\end{equation}
Here, $\mathcal{O}(W)$ is the set of primitives stored in $W$. The system resolves each construction reference by copying its primitives with their original coordinates and geometric parameters. The visual agent controls which objects are shown and highlighted, and the verifier also checks the newly generated objects. For each scene, the fixed renderer produces a keyframe $I_k=\mathcal{R}_{\mathrm{image}}(e_k,\phi_k,\mathcal{O}_k,\mathcal{H}_k)$, yielding $I=(I_1,\ldots,I_K)$.

\subsection{Controlled Scene-to-Video Rendering}
\label{sec:video_stage}

We use a custom scene renderer that draws graphics, instructional text, and formulas according to each scene specification and highlights the designated objects. It produces one keyframe per scene for verification and renders and encodes the video frames in scene order as an MP4 video.

At normalized time $u_n\in[0,1]$, the update operator $\mathcal{M}$ applies predefined motion rules to supported dynamic primitives and leaves static primitives unchanged. The $n$-th frame of scene $k$ is
\begin{equation}
    F_{k,n}=\mathcal{R}_{\mathrm{image}}\!\left(
        e_k,\phi_k,
        \{\mathcal{M}(o,u_n):o\in\mathcal{O}_k\},
        \mathcal{H}_k\right).
    \label{eq:video_frame}
\end{equation}
The renderer holds each scene on screen for reading and concatenates the resulting clips $C_k$ in scene order. It encodes the candidate as $\widetilde V=\operatorname{Encode}(C_1\Vert\cdots\Vert C_K)$. Encoding and decoding checks assess file validity. The verifier uses the associated rendered keyframes for content review.

\subsection{Verification and Refinement}
\label{sec:verification}

The verification agent receives the original problem, execution record, scene specifications, and all rendered keyframes.
\begin{equation}
    (c_m,c_g,c_t,\mathcal{E})
    =\mathcal{A}_{\mathrm{verify}}(q,T,S,I).
    \label{eq:verification_result}
\end{equation}
The agent assesses mathematical correctness ($c_m$), whether the rendered graphics depict the corresponding reasoning steps ($c_g$), and whether the student-facing text is readable ($c_t$). It records defects and their locations in $\mathcal{E}$. Acceptance requires all three binary checks to pass and the issue list to be empty.
\begin{equation}
    a=\mathbf{1}\!\left[c_m=c_g=c_t=1\ \land\ \mathcal{E}=\varnothing\right].
    \label{eq:acceptance}
\end{equation}
A single factual recheck may reassess the reasons for rejection using the same evidence, keyframes, and acceptance criterion.

On rejection, the system sends the reported issues and repair suggestions to the solver agent if $c_m=0$ and to the visual agent otherwise. Revising the solution regenerates $T$ and its dependent scenes, keyframes, and video. Revising the scenes retains $T$ and regenerates the visual outputs. The verifier reviews the regenerated keyframes. Execution, scene validation, and rendering failures return to the stage responsible for the failure.

The loop returns the first candidate that meets the review criteria and allows up to $R=3$ revisions. If no candidate passes within this budget, the system returns the candidate video that best meets the review criteria, together with its review status.

\section{Evaluation}
\label{sec:evaluation}

We use fixed automatic metrics to assess video quality and semantic relevance, and propose a multidimensional instructional evaluation rubric. Based on this rubric, we calibrate the prompt of a multimodal judge using human ratings to assess perceived instructional quality and viewers' self-reported understanding.

\subsection{Video Quality and Semantic Evaluation}
\label{sec:quality_semantic_protocol}

We use imaging quality (IQ), aesthetic quality (AQ), motion smoothness (MS), and text--video similarity (CLIP-T) to assess the image quality, visual aesthetics, motion continuity, and task-related semantic relevance of generated videos.

IQ and AQ use the corresponding image-quality and aesthetic evaluation models from VBench~\cite{huang2024vbench}.
For sampled frames $\{F_i\}_{i=1}^{K}$ from a video $V$, the scores are defined as
\begin{equation}
    \begin{aligned}
        \operatorname{IQ}(V)
        &=\frac{1}{100K}\sum_{i=1}^{K}g_{\mathrm{IQ}}(F_i),\\
        \operatorname{AQ}(V)
        &=\frac{1}{10K}\sum_{i=1}^{K}g_{\mathrm{AQ}}(F_i),
    \end{aligned}
    \label{eq:iq_aq}
\end{equation}
where $g_{\mathrm{IQ}}$ is the MUSIQ-SPAQ image-quality scorer and $g_{\mathrm{AQ}}$ is the aesthetic scorer based on CLIP image features.

MS measures motion smoothness by comparing an intermediate frame predicted from its neighboring frames with the observed frame:
\begin{equation}
    \operatorname{MS}(V)
    =1-\frac{1}{255J}\sum_{j=1}^{J}
    \operatorname{MAE}\!\left(M_j,\mathcal{A}(L_j,R_j)\right),
    \label{eq:motion_smoothness}
\end{equation}
where $(L_j,M_j,R_j)$ is a consecutive frame triplet, $\mathcal{A}$ denotes the AMT-S interpolation model, and $\operatorname{MAE}$ is the mean absolute pixel error.

CLIP-T measures semantic relevance through the similarity between video-frame and task-description features:
\begin{equation}
    \operatorname{CLIP\text{-}T}(V,q)
    =\frac{1}{K}\sum_{i=1}^{K}
    \cos\!\left(f_{\mathrm{img}}(F_i),f_{\mathrm{txt}}(q)\right),
    \label{eq:clip_t}
\end{equation}
where $q$ is the task description, and $f_{\mathrm{img}}$ and $f_{\mathrm{txt}}$ are the CLIP image and text encoders, respectively~\cite{radford2021clip}.

In implementation, IQ, AQ, and CLIP-T use 32 uniformly sampled frames per video, while MS is computed on the original frame sequence; higher scores indicate better performance for all four metrics.

\subsection{Human-Calibrated Instructional Evaluation}
\label{sec:human_calibrated_protocol}

A human-calibrated multimodal judge evaluates videos from the learner's perspective across six dimensions on a five-point scale. Legibility and problem clarity assess readable visual information and clear problem conditions and goals; visual helpfulness and self-reported understanding reflect how graphics support comprehension and viewers' subjective understanding. Step coherence and reasoning clarity assess how solution steps connect and why key steps are taken, emphasizing clear, coherent presentation of the solution process and its rationale in reasoning-intensive videos. Human and judge scores are reported on a 100-point scale ($\times20$).

The human evaluation set comprises 60 generated videos and ten online teaching videos, with 1,577 valid video-level rating records collected from 45 raters. All 45 raters were students taking a data structures and algorithms course. For video $v$ and dimension $d$, we use the mean opinion score across raters~\cite{wang2025aigvassessor} as the human evaluation target:
\begin{equation}
    \mathrm{MOS}_{v,d}
    = \frac{1}{n_{v,d}}\sum_{i=1}^{n_{v,d}}r_{i,v,d},
    \label{eq:human_mos}
\end{equation}
where $r_{i,v,d}$ is the score assigned by rater $i$, and $n_{v,d}$ is the corresponding number of valid ratings.

To align the judge with the human rating scale, we partition the videos by problem family into 38 calibration samples, 12 development samples, and 20 held-out validation samples, with no problem family shared across the sets. Calibration refines the prompt using rubric descriptions, representative human-rated examples, domain statistics, and feedback on prediction errors. The final prompt and judge configuration are selected based on development-set performance. This process does not update model parameters.

The calibrated judge predicts the six human mean scores from the problem statement and video frames, and provides supporting visual evidence. Once frozen, the same judge is applied to videos from all methods and the online teaching references. In implementation, we uniformly sample 24 frames per video and use the same prompt and scoring rules.

\begin{table}[t]
    \centering
    \caption{Multimodal judge agreement with human ratings. Scores and MAE use a 100-point scale.}
    \label{tab:judge_alignment}
    \small
    \setlength{\tabcolsep}{3pt}
    \begin{tabular}{@{}lcccc@{}}
        \hline
        Split & Human & LLM & MAE $\downarrow$ & Spearman $\uparrow$ \\
        \hline
        Calibration      & 90.34 & 90.58 & 4.26 & 0.394   \\
        Validation       & 88.84 & 90.58 & 3.76 & 0.654 \\
        Test             & 89.32 & 90.80 & 4.48 & 0.640 \\
        \hline
    \end{tabular}
\end{table}

\section{Experiments}
\label{sec:experiments}

\begin{table*}[t]
    \centering
    \captionsetup{skip=4pt}
    \caption{\textbf{Video quality and semantic comparison.} Higher is better; \textbf{bold}/\underline{underline} mark best/second-best. Backbone denotes the generation LLM (---: not applicable).}
    \label{tab:quality_semantic_comparison}
    \small
    \fontsize{8.5}{10.2}\selectfont
    \setlength{\tabcolsep}{4pt}
    \renewcommand{\arraystretch}{0.95}
    \begin{tabular}{@{}lllcccc@{}}
        \toprule
        Type & Method & Backbone & IQ $\uparrow$ & AQ $\uparrow$ & MS $\uparrow$ & CLIP-T $\uparrow$ \\
        \midrule
        \multirow{4}{*}{\shortstack[l]{Video generation\\models}}
         & Wan2.1-T2V 1.3B~\cite{wan2025} & --- & \textbf{0.714006} & 0.309320 & 0.987537 & 0.194899 \\
         & LTX-2.5 22B~\cite{Lightricks2026LTX25} & --- & 0.512101 & 0.337238 & 0.991867 & 0.203191 \\
         & Wan2.2-T2V 27B~\cite{wan22t2v} & --- & \underline{0.700417} & 0.352059 & 0.993164 & 0.227803 \\
         & MiniMax H3-33B~\cite{minimax2026h3} & --- & 0.620894 & 0.397049 & 0.993396 & 0.262174 \\
        \midrule
        \multirow{8}{*}{\shortstack[l]{Agent-based\\code generation}}
         & Direct Manim & Qwen3.5-4B & 0.350465 & 0.280453 & \underline{0.997504} & 0.228532 \\
         & Direct Manim & Qwen3.8-27B & 0.410902 & 0.365424 & 0.995869 & \underline{0.271676} \\
         & Code2Video~\cite{chen2026code2video} & Qwen3.5-4B & 0.269708 & 0.306969 & 0.997356 & 0.223261 \\
         & Code2Video~\cite{chen2026code2video} & Qwen3.8-27B & 0.349767 & 0.353048 & \textbf{0.998347} & 0.246425 \\
         & VisualEDU~\cite{chen2025visualedu} & Qwen3.5-4B & 0.384004 & 0.267534 & 0.996292 & 0.171941 \\
         & VisualEDU~\cite{chen2025visualedu} & Qwen3.8-27B & 0.378880 & 0.367361 & 0.995259 & 0.256650 \\
         & ManimAgent~\cite{jiang2026manimagent} & Qwen3.5-4B & 0.343152 & 0.305705 & 0.997420 & 0.239756 \\
         & ManimAgent~\cite{jiang2026manimagent} & Qwen3.8-27B & 0.365310 & 0.351190 & 0.997301 & 0.260913 \\
        \midrule
        \multirow{2}{*}{Our method}
         & \textbf{ReVAgent} & Qwen3.5-4B & 0.619047 & \underline{0.462111} & 0.996861 & 0.233369 \\
         & \textbf{ReVAgent} & Qwen3.8-27B & 0.691707 & \textbf{0.577807} & 0.996936 & \textbf{0.273723} \\
        \bottomrule
    \end{tabular}
\end{table*}

\begin{table*}[!t]
    \centering
    \captionsetup{skip=4pt}
    \caption{\textbf{Human-calibrated instructional comparison.} Frozen judge predictions on a 100-point scale. Higher is better; \textbf{bold}/\underline{underline} mark best/second-best among generation methods.}
    \label{tab:human_calibrated_comparison}
    \small
    \fontsize{8.5}{10.2}\selectfont
    \setlength{\tabcolsep}{3pt}
    \renewcommand{\arraystretch}{0.95}
    \begin{tabular}{@{}lllcccccc@{}}
        \toprule
        Type & Method & Backbone
        & Legibility $\uparrow$
        & \shortstack{Problem\\Clarity $\uparrow$}
        & \shortstack{Step\\Coherence $\uparrow$}
        & \shortstack{Reasoning\\Clarity $\uparrow$}
        & \shortstack{Visual\\Helpfulness $\uparrow$}
        & \shortstack{Self-reported\\Understanding $\uparrow$} \\
        \midrule
        \multirow{4}{*}{\shortstack[l]{Video generation\\models}}
         & Wan2.1-T2V 1.3B~\cite{wan2025} & --- & 25.60 & 22.04 & 21.84 & 20.54 & 22.62 & 21.22 \\
         & LTX-2.5 22B~\cite{Lightricks2026LTX25} & --- & 38.32 & 36.42 & 30.42 & 25.98 & 31.12 & 27.88 \\
         & Wan2.2-T2V 27B~\cite{wan22t2v} & --- & 49.78 & 38.28 & 37.48 & 30.04 & 39.48 & 32.76 \\
         & MiniMax H3-33B~\cite{minimax2026h3} & --- & 55.60 & 59.22 & 49.00 & 40.66 & 47.36 & 47.12 \\
        \midrule
        \multirow{8}{*}{\shortstack[l]{Agent-based\\code generation}}
         & Direct Manim & Qwen3.5-4B & 76.33 & 72.67 & 58.33 & 54.00 & 56.00 & 59.00 \\
         & Direct Manim & Qwen3.8-27B & 89.92 & 90.90 & 86.78 & 82.10 & 83.56 & 84.96 \\
         & Code2Video~\cite{chen2026code2video} & Qwen3.5-4B & 81.33 & 85.33 & 71.67 & 69.25 & 65.75 & 72.33 \\
         & Code2Video~\cite{chen2026code2video} & Qwen3.8-27B & 90.70 & 91.40 & 86.56 & 82.48 & \underline{84.24} & 85.12 \\
         & VisualEDU~\cite{chen2025visualedu} & Qwen3.5-4B & 84.00 & 90.00 & 76.00 & 72.00 & 68.00 & 78.00 \\
         & VisualEDU~\cite{chen2025visualedu} & Qwen3.8-27B & 89.48 & 88.32 & 83.02 & 76.60 & 80.54 & 80.98 \\
         & ManimAgent~\cite{jiang2026manimagent} & Qwen3.5-4B & 74.38 & 71.28 & 63.83 & 59.88 & 57.85 & 63.45 \\
         & ManimAgent~\cite{jiang2026manimagent} & Qwen3.8-27B & 90.68 & 88.50 & 83.04 & 78.30 & 81.60 & 81.96 \\
        \midrule
        \multirow{2}{*}{Our method}
         & \textbf{ReVAgent} & Qwen3.5-4B & \underline{93.62} & \underline{92.15} & \underline{89.23} & \underline{86.15} & 82.31 & \underline{87.62} \\
         & \textbf{ReVAgent} & Qwen3.8-27B & \textbf{95.32} & \textbf{93.64} & \textbf{92.12} & \textbf{88.40} & \textbf{90.10} & \textbf{89.56} \\
        \bottomrule
    \end{tabular}
\end{table*}

We evaluate ReVAgent on mathematical and algorithmic instructional video generation tasks. Combining video quality and semantic evaluation with human-calibrated instructional evaluation, we compare the instructional quality, visual quality, and semantic relevance of videos generated by different methods. We also use qualitative examples to analyze reasoning steps and visual explanations, and ablation studies to examine the roles of the solver, visual, and verification agents.

\subsection{Experimental Setup}
\label{sec:experimental_setup}

The planned comparison uses a shared set of 100 tasks, including 50 mathematical and 50 algorithmic tasks. We compare ReVAgent with four video generation models, Wan2.1-T2V 1.3B, Wan2.2-T2V 27B~\cite{wan22t2v}, MiniMax H3-33B, and LTX-2.5 22B~\cite{Lightricks2026LTX25}, as well as Direct Manim, Code2Video, VisualEDU, and ManimAgent. Code-based methods and ReVAgent use Qwen3.8-27B and Qwen3.5-4B backbones. On shared tasks, methods receive the same problems, input data, and instructional requirements. The four text-to-video models use the same fixed 100 English-language tasks. All methods are evaluated with the same automatic metrics and frozen human-calibrated judge.

\subsection{Video Quality and Semantic Comparison}
\label{sec:quality_semantic_comparison}

Among the video samples reported in Table~\ref{tab:quality_semantic_comparison}, the 27B ReVAgent configuration shows its largest lead over the baselines in aesthetic quality (AQ), scoring 0.578 and exceeding the strongest baseline, MiniMax H3-33B, by 0.181. ReVAgent also ranks first in task-related semantic relevance (CLIP-T), edging out Direct Manim, the strongest code-based baseline on this metric, by 0.002. Both methods use code to explicitly specify and render formulas, diagrams, and labels. This can help preserve task-specific content that direct pixel generation may distort or omit. The pixel-based baselines in the table obtain lower CLIP-T scores.

Wan2.1 and Wan2.2 achieve the two highest imaging quality (IQ) scores, illustrating the advantage pixel-based generators can have in low-level frame quality. ReVAgent remains close to both, trailing them by only 0.022 and 0.009, respectively. On motion smoothness (MS), ReVAgent trails the best-scoring method, Code2Video, by only 0.0014. Section~\ref{sec:human_calibrated_comparison} further examines formula legibility, the coherence of reasoning steps, and how well diagrams support the explanation.

\subsection{Human-Calibrated Instructional Comparison}
\label{sec:human_calibrated_comparison}

Table~\ref{tab:human_calibrated_comparison} shows that the 27B ReVAgent configuration achieves the highest scores across all six evaluation dimensions in the reported video subsets, with a six-dimensional average of 91.52. Its step coherence, reasoning clarity, and visual helpfulness scores reach 92.12, 88.40, and 90.10, exceeding Code2Video by 5.56, 5.92, and 5.86 points, respectively. These results indicate favorable organization of reasoning steps, presentation of supporting explanations, and correspondence between diagrams and instructional content within the evaluated samples.

Wan2.1 and Wan2.2 have the two highest reported IQ values and six-dimensional judge means of 22.30 and 37.98, respectively. MiniMax H3-33B and LTX-2.5 22B obtain six-dimensional means of 49.83 and 31.69, respectively, both well below ReVAgent's 91.52.

Although some code-based baselines score close to ReVAgent on legibility and problem clarity, their rendered outputs still differ in how they position labels relative to diagrams and organize content on screen. The next section uses a common Heron's formula task to examine how these differences affect the reader's ability to follow the derivation.

\begin{figure*}[t]
    \centering
    \newlength{\heronimagewidth}
    \newlength{\heronimageheight}
    \newcommand{\heronpanel}[2]{%
        \begin{subfigure}[t]{0.328\textwidth}
            \phantomsubcaption
            \setlength{\heronimagewidth}{\dimexpr\linewidth-12pt\relax}%
            \setlength{\heronimageheight}{0.5625\heronimagewidth}%
            \noindent\hbox{%
                \vbox to\heronimageheight{\vfil%
                    \hbox to10pt{\hss\rotatebox{90}{\fontsize{8.5}{9.5}\selectfont(\thesubfigure)~#2}\hss}%
                    \vfil}%
                \kern2pt%
                \vbox to\heronimageheight{\vfil%
                    \hbox to\heronimagewidth{\hss%
                        #1%
                        \hss}%
                    \vfil}%
            }%
        \end{subfigure}%
    }
    \noindent\heronpanel{\includegraphics[width=\heronimagewidth,height=\heronimageheight,keepaspectratio]{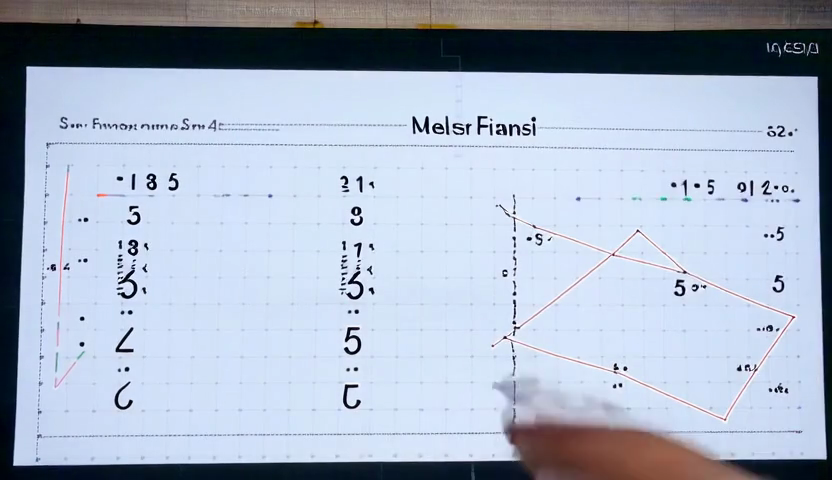}}{Wan2.1-T2V 1.3B}\hfill%
    \heronpanel{\includegraphics[width=\heronimagewidth,height=\heronimageheight,keepaspectratio]{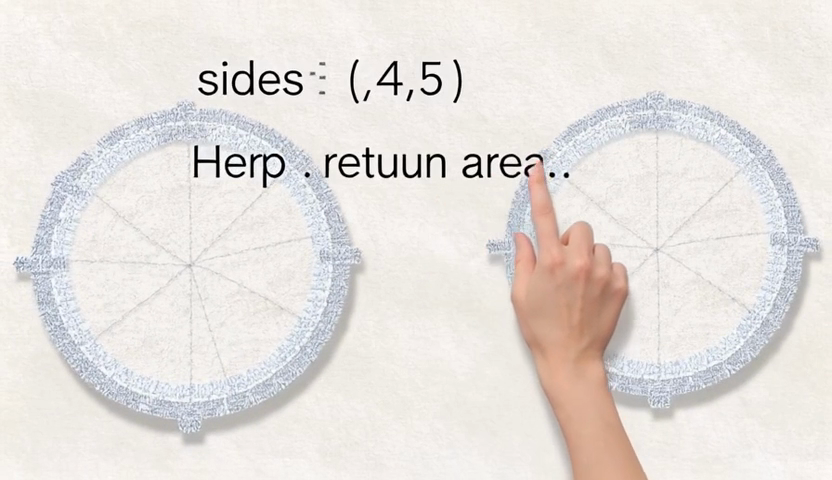}}{LTX-2.5 22B}\hfill%
    \heronpanel{\includegraphics[width=\heronimagewidth,height=\heronimageheight,keepaspectratio]{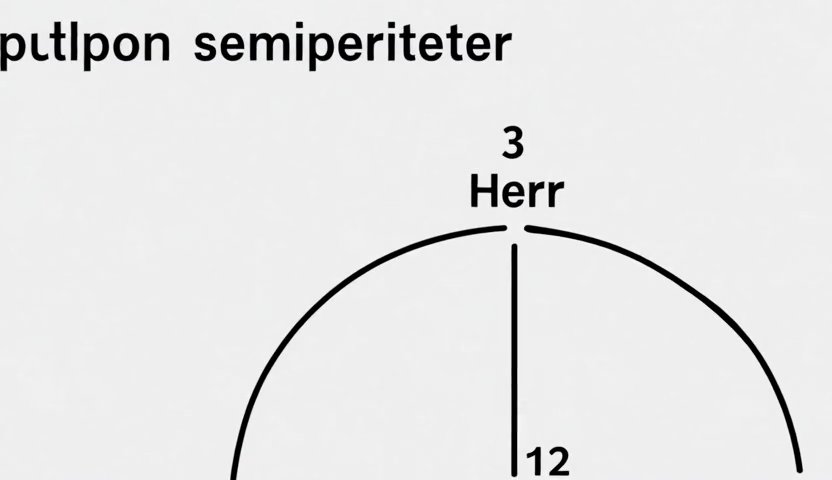}}{Wan2.2-T2V 27B}\par
    \nointerlineskip\vspace{2pt}
    \noindent\heronpanel{\includegraphics[width=\heronimagewidth,height=\heronimageheight,keepaspectratio]{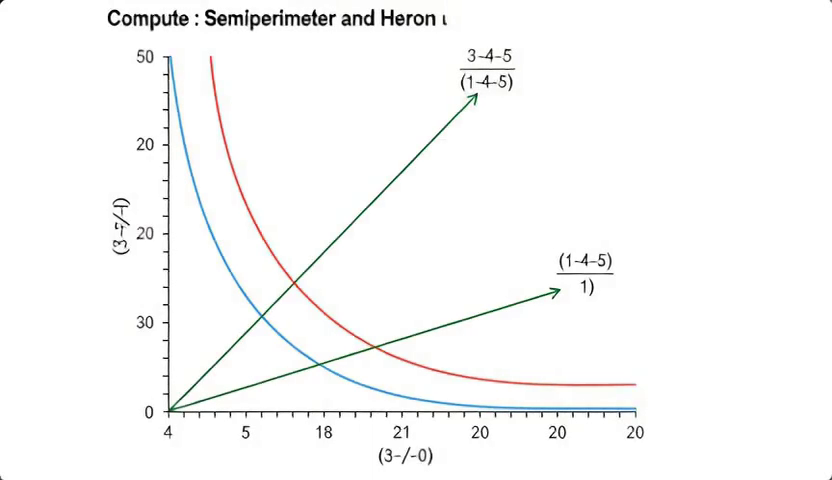}}{MiniMax H3-33B}\hfill%
    \heronpanel{\includegraphics[width=\heronimagewidth,height=\heronimageheight,keepaspectratio]{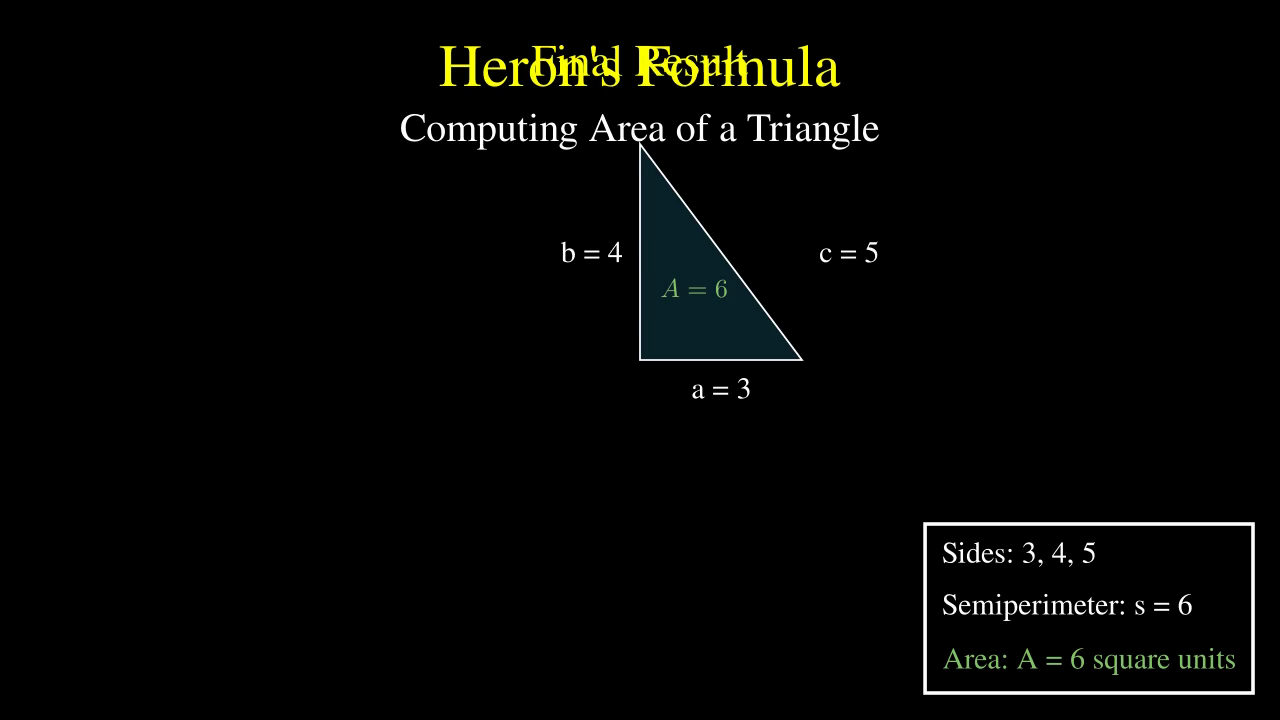}}{Direct Manim}\hfill%
    \heronpanel{\includegraphics[width=\heronimagewidth,height=\heronimageheight,keepaspectratio]{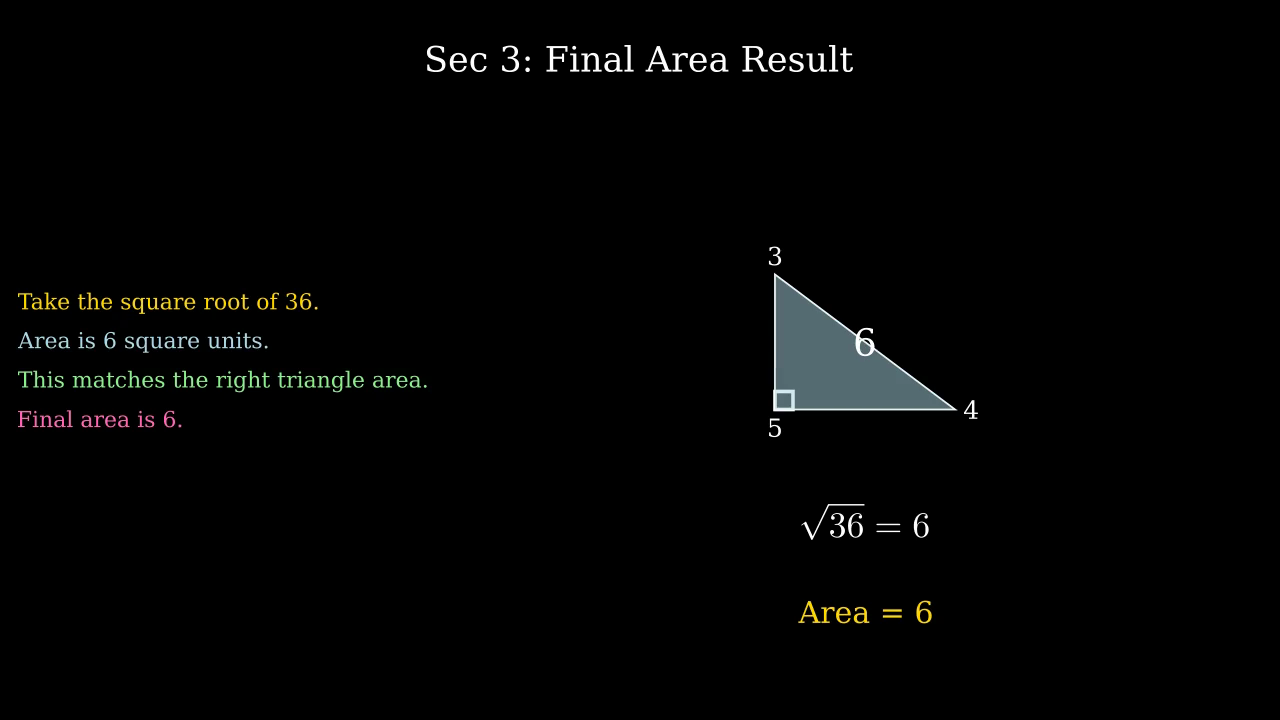}}{Code2Video}\par
    \nointerlineskip\vspace{2pt}
    \noindent\heronpanel{\includegraphics[width=\heronimagewidth,height=\heronimageheight,keepaspectratio]{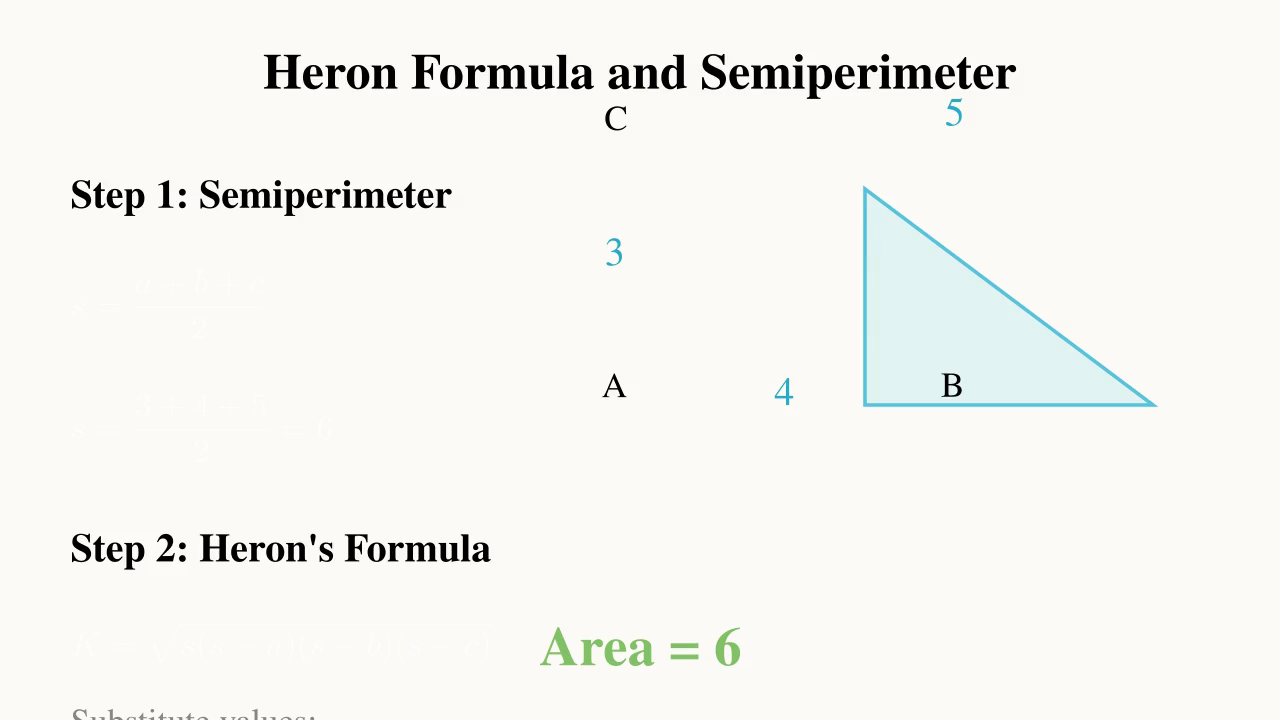}}{VisualEDU}\hfill%
    \heronpanel{\includegraphics[width=\heronimagewidth,height=\heronimageheight,keepaspectratio]{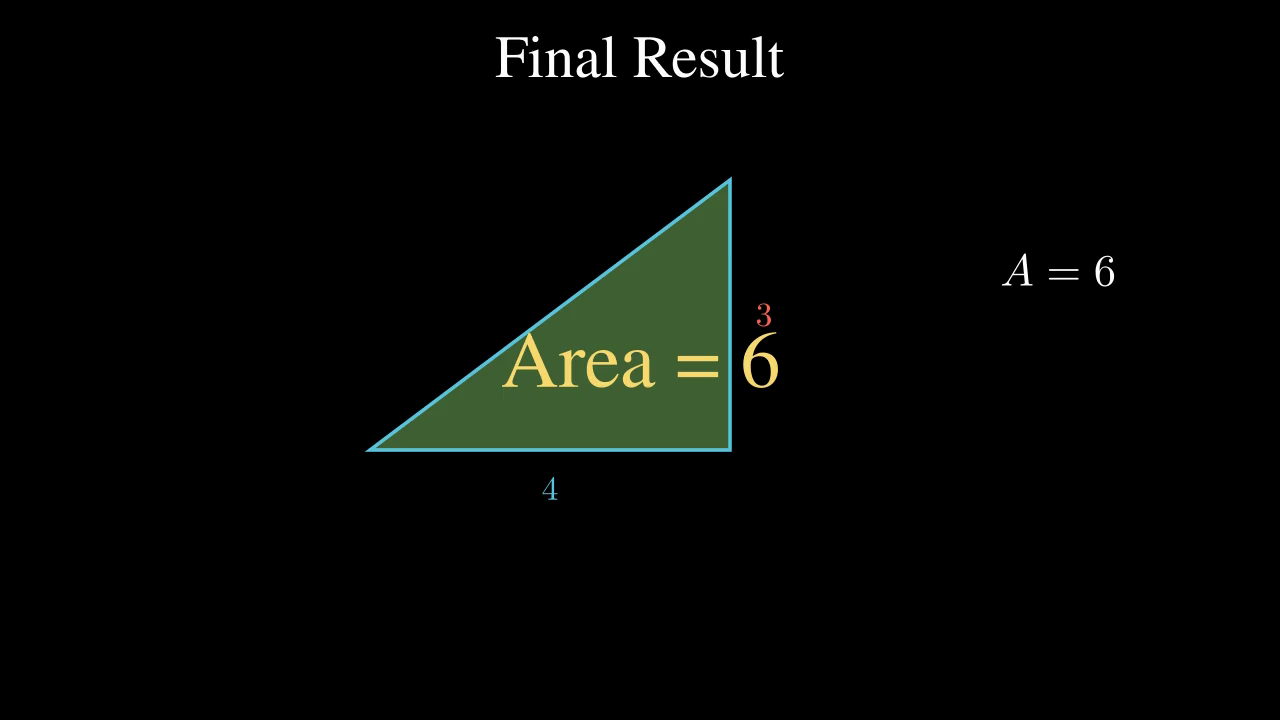}}{ManimAgent}\hfill%
    \heronpanel{\includegraphics[width=\heronimagewidth,height=\heronimageheight,keepaspectratio]{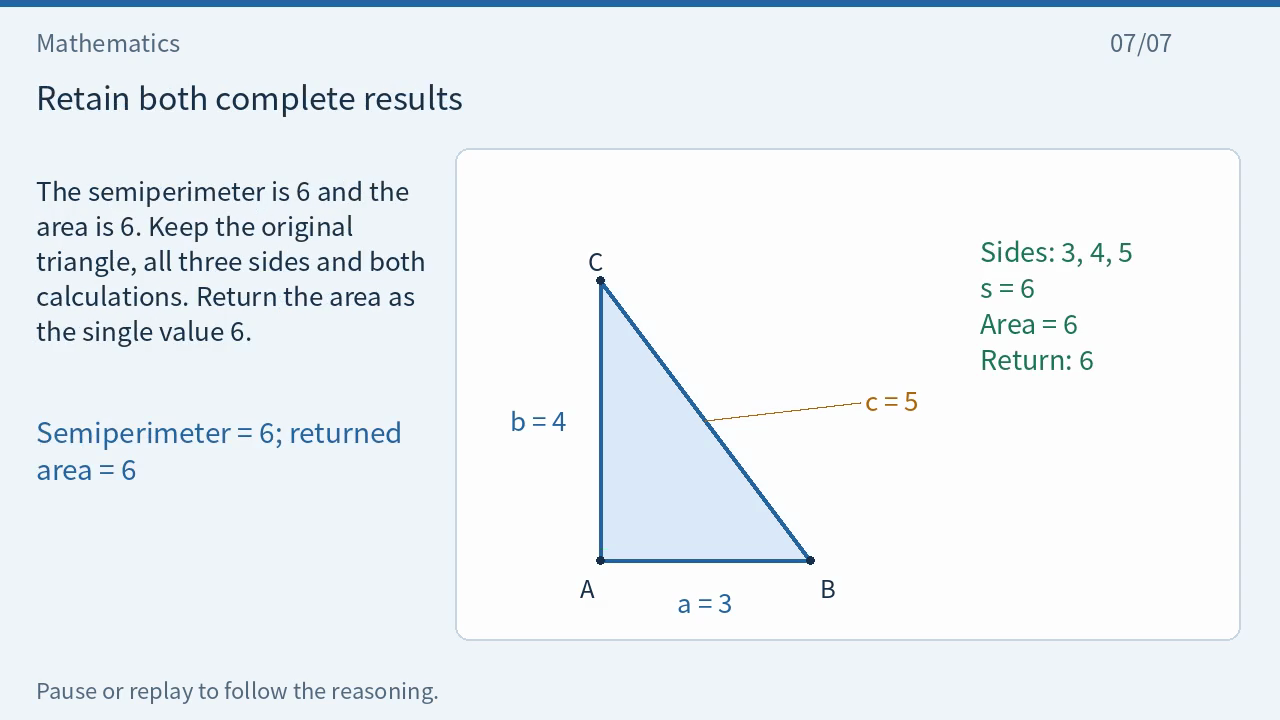}}{\textbf{ReVAgent (Ours)}}\par

    \caption{\textbf{Comparison of visualization results.} All nine methods address the same Heron's formula task with side lengths 3, 4, and 5. The panels show full frames at 95\% of each video. Panels (a--d) show video generation models, (e--h) show code-based baselines, and (i) shows ReVAgent.}
    \label{fig:qualitative_comparison}
\end{figure*}

\subsection{Visualization Analysis}
\label{sec:visualization_analysis}

Figure~\ref{fig:qualitative_comparison} compares nine methods on the same Heron's formula task at 95\% video progress. The pixel-based text-to-video models exhibit errors in mathematical notation and geometry. Wan2.1 and Wan2.2 obtain the two highest imaging quality (IQ) scores, yet some text and numbers in the Wan2.1 frame are difficult to read, and Wan2.2 misspells the semiperimeter label and depicts a curved outline in place of the triangle. MiniMax H3-33B renders unrelated coordinate curves with malformed labels. LTX-2.5 22B shows two circular outlines and an incomplete list of side lengths. These errors make it difficult to connect the visuals to the problem statement and the calculation steps.

The code-based baselines show the correct area, but some labels and layouts make the frames harder to read. Code2Video places the side lengths near the vertices, leaving viewers to infer which length belongs to each edge. Direct Manim has overlapping titles, while VisualEDU places some labels away from the corresponding diagram elements, displays low-contrast formulas, and clips content at the bottom of the frame. For beginners, these issues make it harder to interpret the labels and follow the derivation, reducing the videos' suitability as standalone teaching material for a wider audience.

ReVAgent presents the explanation and diagram in separate areas of the frame and retains the original triangle, all three side labels, the semiperimeter, and the area in the final view.
\ifdefined\arxivwithoutappendix\else
Intermediate frames in the supplementary material show that the same triangle and labels remain visible as the video substitutes values into Heron's formula and evaluates the square root, allowing viewers to follow the calculation against a consistent diagram.
\fi
The differences in layout, contrast, and label placement are consistent with the aesthetic quality (AQ) results in Sec.~\ref{sec:quality_semantic_comparison}, where ReVAgent scores 0.578 and the best code-based baseline scores 0.367.

\subsection{Generation Efficiency}
\label{sec:generation_efficiency}

\begin{figure*}[!t]
    \centering
    \includegraphics[width=\textwidth]{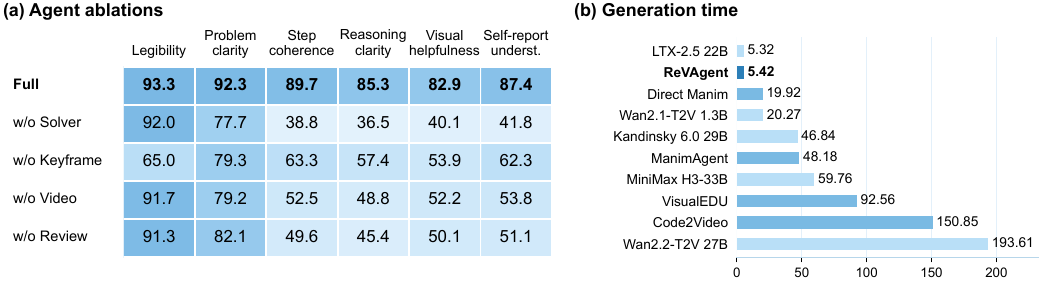}
    \captionsetup{skip=3pt}
    \caption{\textbf{Agent ablations and generation time.} (a) Six-dimensional quality scores with Qwen3.5-4B. (b) Generation time analysis.}
    \label{fig:ablation_efficiency}
    \label{fig:generation_efficiency}
\end{figure*}

We measure generation efficiency as generation time per minute of output video, computed by dividing total generation time by total actual playback duration. Timing covers inference, rendering or encoding, and the method's own revisions. ReVAgent takes 5.42 minutes per minute of video, compared with 19.92 minutes for Direct Manim, or about 27\% of its time. VisualEDU and ManimAgent take 92.56 and 48.18 minutes, respectively. ReVAgent has a lower generation time per minute of video than all four Manim baselines (Figure~\ref{fig:ablation_efficiency}(b)). Our method reuses a fixed renderer and organizes execution results into structured scenes, avoiding the need to generate and debug a complete animation program for each problem. LTX-2.5 22B takes 5.32 minutes per minute of video, while its six instructional quality scores range from 25.98 to 38.32, with low step coherence and reasoning clarity.

The qualitative analysis and quality evaluation show that longer generation time does not necessarily lead to better instructional presentation. Wan2.2 and MiniMax H3-33B take longer to generate videos than ReVAgent, yet their videos still exhibit errors in mathematical notation and diagrams that deviate from the problem. LTX-2.5 22B is slightly faster, but its low step coherence and reasoning clarity indicate that fast visual generation alone is insufficient for a clear instructional explanation. ReVAgent achieves higher instructional quality scores while maintaining a low generation cost, providing a better balance between generation efficiency and instructional quality.

\subsection{Agent Ablations}
\label{sec:agent_ablations}

We ablate four agents on the same 20 tasks with Qwen3.5-4B (Figure~\ref{fig:ablation_efficiency}(a)). Fixed reference solutions replace the solver; Qwen-Image generates keyframes from the problem and steps; RIFE interpolates the original keyframes; and removing the reviewer disables visual review and feedback. Full leads all six dimensions. Removing the solver and reviewer lowers reasoning clarity by 57.2\% and 46.8\%, respectively. Replacing the keyframe agent lowers legibility by 30.3\%. RIFE retains 91.7 legibility but 38.4\% lower self-reported understanding, suggesting that smooth interpolation alone cannot preserve reasoning continuity.

\begingroup
\setlength{\intextsep}{3pt}
\begin{table}[!ht]
    \centering
    \captionsetup{skip=3pt}
    \caption{\textbf{Reasoning--visual alignment.} Complete: valid format/rendering and correct four-stage geometry/state with original labels.}
    \label{tab:execution_record_mechanisms}
    \small
    \setlength{\tabcolsep}{4pt}
    \renewcommand{\arraystretch}{1.0}
    \begin{tabular}{@{}lcc@{}}
        \toprule
        Condition & Complete $\uparrow$ & Mean output tokens $\downarrow$ \\
        \midrule
        Use per-step graphics & 20/20 & 1033 \\
        Rewrite drawing commands & 15/20 & 2981 \\
        No diagram--step links & 11/20 & 1156 \\
        Final answer + input figure & 3/20 & 2685 \\
        \bottomrule
    \end{tabular}
\end{table}

To examine errors introduced when correct solutions become instructional graphics, we use 20 verified programmatic solutions and their per-step construction coordinates in a separate controlled study with Qwen3.5-4B, without review. Reuse and rewriting receive identical data; construction catalogs are shuffled. Direct use omits fewer input labels and reduces output tokens by 65.3\% (Table~\ref{tab:execution_record_mechanisms}). Removing construction--step links or supplying only the final answer and input figure makes intermediate stages less consistent. These results support our cross-agent reasoning--visual alignment: reusing the solver's constructions at the corresponding explanation steps reduces omissions and stage mismatches. This criterion directly checks the correctness of graphics and states at each stage, and the results support our method's ability to preserve consistency between intermediate reasoning states and their visual representations in reasoning-intensive instructional video generation.
\endgroup

\section{Conclusion}
\label{sec:conclusion}

We introduced ReVAgent, a multi-agent framework for generating instructional
videos for mathematics and algorithms. ReVAgent uses a shared program
execution record to build instructional scenes and verify rendered keyframes,
linking formulas, diagrams, and state changes to specific solution steps.
We also developed a six-dimensional rubric for instructional quality and used
human ratings to calibrate a multimodal judge. On the evaluated video samples,
ReVAgent achieved the highest aesthetic quality score and ranked first among
the compared generation methods on all six dimensions assessed by the judge,
including step coherence, reasoning clarity, and visual helpfulness.
Executable reasoning thus supports coherent visual explanations.

\section*{LLM Usage Statement}

In our experiments, ReVAgent uses the models identified as Qwen3.8-27B and Qwen3.5-4B for
problem solving, scene planning, and keyframe verification. The code-generation
baselines are evaluated with the same backbones, while the video generation baselines use their
respective pretrained models. For human-calibrated instructional evaluation, we calibrate a prompt
for the frozen Qwen3.8-27B judge using human ratings, then apply the fixed prompt to
score videos on six dimensions of instructional quality.

We use OpenAI Codex to assist with manuscript writing, literature search,
citation verification, and code development.
The code supports running experiments, aggregating results, revising selected
diagrams, and adjusting \LaTeX{} layout. The authors take responsibility for all
submitted content.

\end{document}